\documentclass[pdflatex,sn-mathphys-num]{sn-jnl}

\usepackage{graphicx}%
\usepackage{multirow}%
\usepackage{amsmath,amssymb,amsfonts}%
\usepackage{amsthm}%
\usepackage{mathrsfs}%
\usepackage[title]{appendix}%
\usepackage{xcolor}%
\usepackage{textcomp}%
\usepackage{manyfoot}%
\usepackage{booktabs}%
\usepackage{url}
\usepackage{hyperref}
\usepackage{algorithm}%
\usepackage{algorithmicx}%
\usepackage{algpseudocode}%
\usepackage{listings}%

\theoremstyle{thmstyleone}%
\theoremstyle{thmstyletwo}%

\theoremstyle{thmstylethree}%

\begin{document}

\title[Article Title]{Sensing to Intelligence:
Principles for Neuromorphic Circuits and Systems}


\author[1,2]{\fnm{Saptarshi} \sur{Maiti}}\email{saptarshimai@iisc.ac.in}

\author*[1,2]{\fnm{Chetan Singh} \sur{Thakur}}\email{csthakur@iisc.ac.in}


\affil[1]{\orgdiv{Department of Electronic Systems Engineering}, \orgname{Indian Institute of Science}, \orgaddress{\city{Bangalore}, \postcode{560012}, \country{India}}}

\affil[2]{\orgdiv{Brain, Computation, and Data Science}, \orgname{Indian Institute of Science}, \orgaddress{\city{Bangalore}, \postcode{560012},  \country{India}}}


\abstract{Neuromorphic engineering began with the idea that the physical behavior of a system could itself be used for computation, taking inspiration from the way nervous systems sense, adapt, and evolve in time. The field has since expanded far beyond its early analog circuits to include event-based sensors, spiking processors, emerging memory devices, mixed-signal systems, and large-scale neural accelerators. With this expansion, however, the meaning of \textit{neuromorphic} has become increasingly broad. In this Perspective, we argue that neuromorphic engineering should be defined neither by resemblance to biological components nor by any particular device, signal representation, or substrate. Its potential lies in identifying computational principles in biological systems and translating them into the organization and dynamics of artificial machines. A system is truly neuromorphic when the invoked biological or physical principle plays a causal, design-relevant role in how information is represented, how state evolves, or what capability the complete system achieves. We develop this view across sensing, collective computation, memory, learning, and interaction. Adaptation, nonlinear dynamics, attractor structure, variability, and closed-loop action illustrate how computation can be embedded in a machine's evolving physical state. From these examples, we identify a set of design principles for neuromorphic systems. As neuromorphic engineering moves toward wider deployment, this perspective shifts the central question from how closely machines resemble nervous systems to what computation becomes possible when their principles are understood and deliberately translated into new machines.
}

\maketitle
\section{The Neuromorphic Origin}
\label{sec1}
The brain has been in an unusual position in science: it is the system through which we perceive the world, yet it remains among the systems we understand least completely. Shaped over millions of years of evolution, nervous systems achieve perception, memory, learning, and interaction with extraordinary efficiency that engineering still cannot match. From Ramón y Cajal's neuron doctrine in the late nineteenth century to Hodgkin and Huxley's quantitative description of neural signaling in the 1950s, neuroscience revealed many of the brain's fundamental building blocks \cite{LopezMunoz2006Neuron, HodgkinHuxley1952}. Yet, more than a century later, we still lack a general understanding of how these elements collectively give rise to intelligence \cite{Roland2023HowFar}.

At the same time, this growing understanding of the nervous system inspired the idea of building some of its computational principles into machines. One of the pioneers of modern digital electronics, Carver Mead, began exploring this possibility in the late 1960s after molecular biologist Max Delbrück introduced him to the similarities in electrical behavior of neurons and transistors, drawing Mead toward the biophysics of neural computation \cite{DelbruckLiuMead2026}. In the following decade, Eric Vittoz and colleagues showed that Metal-Oxide-Semiconductor (MOS) transistors operating in weak inversion could enable extremely low-power analog computation \cite{VittozFellrath1977}. Together, these developments suggested a different way of thinking about silicon. Rather than treating transistors only as digital switches, their nonlinear device physics could itself participate in computation, similar to membranes, ion channels, and synapses in nervous systems.

These threads came together at Caltech in the early 1980s through an unusual convergence of physics, engineering, and neuroscience. Mead's discussions with John Hopfield and Richard Feynman led to them jointly teaching the interdisciplinary course, \textit{The Physics of Computation} during 1981-83 \cite{feynman2018feynman,mead2022oralhistory}, at the same time when Hopfield was showing how collective recurrent dynamics could store memories as stable attractors \cite{hopfield1982neural}. Mead's growing interactions with neuroscience brought another decisive figure: Misha Mahowald, a biology graduate who joined Mead as a PhD student and proposed a remarkably direct experiment: ``Let's build a retina." \cite{DelbruckLiuMead2026}. Mead would later credit her with leading him to the \textit{neuromorphic way of thinking} \cite{mead2023mahowald}. What followed was an extraordinary period of exploration: silicon retinas, cochleae, neurons, winner-take-all circuits, and eventually, event-based communication emerged \cite{mead1988silicon,mahowald1992vlsi,mahowald1994stereoscopic}. The objective was not just to reproduce every biological detail, but to discover the principles of nervous systems that could become principles of engineered systems. Mead crystallized this philosophy in 1990 under the name \textit{neuromorphic electronic systems} \cite{mead1990neuromorphic}. The idea went beyond building electronic neurons. He argued that biological information-processing systems follow principles fundamentally different from those of conventional computers, and that silicon offered the possibility of directly implementing some of those principles in physical devices through its behavior and organization \cite{mead1989analog}.

However, the ambition to build brain-inspired computing systems included earlier contributions that were often overshadowed as later narratives took shape. Back in the 1970s, Shun-ichi Amari showed learned patterns could form stable states in recurrent networks, Kaoru Nakano introduced the Associatron for recalling complete patterns from partial cues, and William Little described persistent collective states in recurrent binary networks \cite{Amari1972LearningPatterns, Nakano1972Associatron, Little1974PersistentStates}-- nearly a decade before Hopfield’s influential formulation of associative memory. Amari’s work on adaptive learning and Kunihiko Fukushima’s multilayer networks anticipated the ideas that later became central to modern neural networks \cite{Amari1967AdaptivePattern, Fukushima1969VisualFeature, Fukushima1980Neocognitron}. Many foundations of neural computation emerged through parallel, sometimes under-recognized works before related ideas resurfaced in new formulations and, with new terminology and computational resources, gained broader recognition.

Nevertheless, Caltech brought together physics, neural dynamics, and biological organization into a distinct engineering philosophy. \textit{The Physics of Computation} course captured a deeper idea-- computation is a physical process, and when many simple elements interact, collective phenomena can become computational resources \cite{feynman2018feynman}. Through this convergence, neuromorphic engineering emerged as a way of treating physical dynamics as computational resources. This history leaves a sharper question: what should it mean, in engineering terms, for a machine to compute like a nervous system?

\begin{figure}[t]
    \centering
    \includegraphics[width=0.8\textwidth]{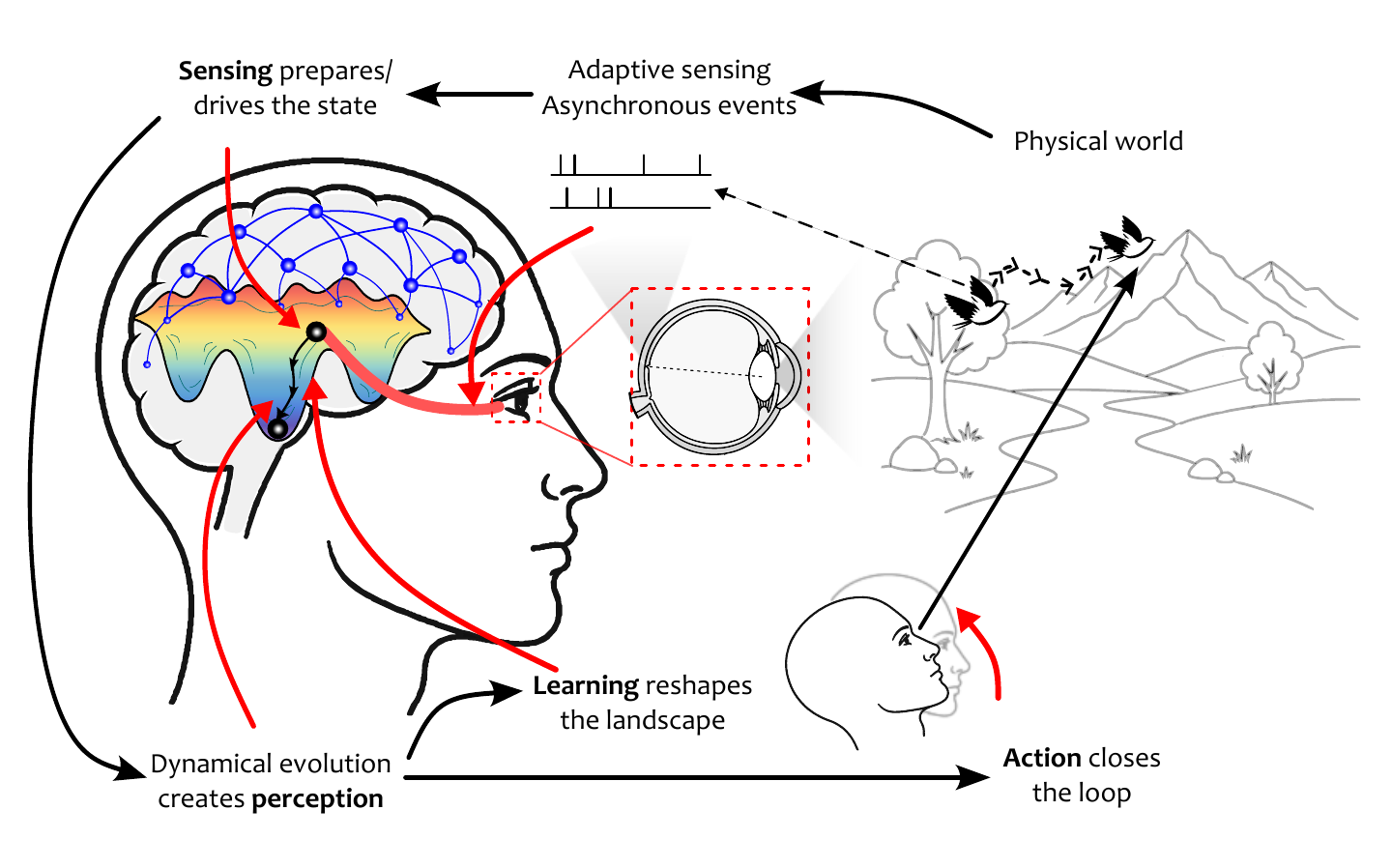}
    \caption{\textbf{Dynamical blueprint for neuromorphic systems.} The physical world provides continuous sensory input (right). The eye illustrates adaptive sensing (center): the scene is projected onto the retina, generating asynchronous spikes in response to local temporal changes-- the flying bird produces sparse activity while the static background is largely suppressed. Sensing performs selection and compression before information reaches the brain. In the brain, these events establish and continuously drive the collective state of a recurrent neural network. Its interactions shape a dynamical landscape, illustrated as an energy landscape (left), through which network activity evolves toward stable states, or attractors, that underlie perception. Over longer timescales, learning modifies these interactions and reshapes the landscape, altering the trajectories available during future perception. Perception then guides action-- for example, orienting toward the bird (bottom-right)-- which changes the sensory input for the next cycle and closes the loop with the environment. Together, these stages form a sensorimotor loop in which sensing, recurrent dynamics, learning, and action jointly shape the system's future state.}
    \label{fig:dynamic}
\end{figure}

This Perspective returns to that question by developing a system-level view in which sensing creates internal state, collective dynamics organize its evolution into perception, and learning reshapes future trajectories (Fig.~\ref{fig:dynamic}). The goal is not to reproduce biology component by component, but to identify the broader principles of neural computation that can guide the design of neuromorphic systems.

\section{Neuromorphic Beyond the Substrate}
Neuromorphic engineering began with the insight that the physical substrate can participate directly in computation. Early analog VLSI made this idea concrete: transistor nonlinearities and dynamics could directly perform useful computation. However, a computational principle should persist regardless of physical realizations. A state may be stored as charge on a capacitor, as conductance in a memory device, or as a digital variable; recurrence may arise through analog currents or stored digital connectivity; delay may arise from physical propagation or a programmable counter-- the implementation is secondary to the computational role. The substrate matters because it determines which particular dynamics are natural, efficient, robust, or scalable to realize. But the neuromorphic claim should attach to the organization of state and interaction. A system is not intrinsically neuromorphic because it is analog, digital, CMOS, memristive, photonic, or tied to any other particular substrate.

Biology itself mixes continuous state, discrete events, nonlinear transformations, delays, and communication. Silicon, for example, can realize the same computational roles in different ways. Neurogrid and BrainScaleS-2 combine analog dynamics with digital communication, whereas SpiNNaker and Loihi implement neuromorphic organization largely through digital many-core systems \cite{Benjamin2014Neurogrid, Pehle2022BrainScaleS2, Furber2014SpiNNaker, Davies2018Loihi}. The implementations differ; what matters is the role that the state and its evolution play in the complete system. This extends beyond dedicated neuromorphic processors. Field programmable gate arrays (FPGAs) and communication fabrics can also serve as substrates. The recently proposed Processing-in-Interconnect leverages delays inherent in commercial routers and switches to perform computation \cite{Srivatsav2025Pi2}. Communication then does not simply move state; its timing and delay dynamics directly participate in the computation.

Biology and silicon differ profoundly in physical organization. Neural tissue is three-dimensional: neurons and synapses occupy and connect through the same volume. More than 200,000 cells and about half a billion synapses were reconstructed from roughly a cubic millimeter of mouse visual cortex \cite{MICrONS2025}. Modern transistors are even smaller than neurons, but the challenge lies in connectivity-- silicon is not equivalent to neural tissue. Nor should it be. Nervous systems evolved under constraints of ionic signaling, metabolism, growth, and biological timescales. Silicon inherits a different set of advantages-- CMOS switching is orders of magnitude faster than neural timescales, digital state can be reliably copied and routed, and programmable logic can provide precision and flexibility that biology lacks. Biological timescale, precision, topology, or device physics should therefore not be treated as specifications that neuromorphic hardware must reproduce.

Biology should therefore provide principles, not hardware specifications. Hardware may realize those principles through entirely different devices, speeds, and physical organizations. The substrate does not make a system neuromorphic; the way its state, interactions, and dynamics are organized does.

\section{What Makes a System Neuromorphic?}

As neuromorphic engineering has expanded across neuroscience,  circuits, architectures, devices, and machine learning, the term itself has become increasingly broad \cite{christensen2022roadmap,kudithipudi2025scale,schuman2022opportunities}. A neuron-like device, an event-based processor, an in-memory accelerator, and a large spiking system may all be called neuromorphic, but they show biological inspiration at very different levels. This diversity has advanced the field, but it also raises the question: what should make a system neuromorphic?

Biological resemblance alone is not enough. A spike used solely as a compact data packet differs from an event whose timing participates in recurrent dynamics; a programmable conductance may resemble a synapse while the surrounding computation remains essentially conventional. As the field has grown, biological language has sometimes been applied to familiar computational units rather than biology being used to challenge computational abstractions themselves. Many such developments have been motivated by genuine properties of nervous systems, including low-energy operation, distributed state, and sparse event-driven communication. The limitation arises when these ideas are reduced to isolated primitives--neurons, synapses, weights, or spikes--without preserving the organization that gives them their computational role.

It is useful to distinguish three levels of inspiration. First, \textbf{component imitation} reproduces selected properties of neurons, synapses, or sensory receptors. Second, \textbf{operation acceleration} uses specialized hardware to execute an established computation more efficiently. Third, \textbf{principle-driven system design} asks what capability a biological mechanism provides, what organization and dynamics make that capability possible, and how an artificial system can realize the same principle. The first two levels can support neuromorphic engineering, but the third defines its central opportunity. A system is not strongly neuromorphic simply because it uses spikes, stores weights in emerging memory devices, accelerates neural network operations, or labels its devices as synapses. Neuromorphic engineering begins when these elements are organized around biological principles of computation at the system level.

Neuromorphic progress ultimately has to be evaluated at the level of the complete system. Emerging memories and crossbar arrays can store parameters and perform matrix-vector multiplication with high parallelism and energy efficiency \cite{sebastian2020memory,aguirre2024hardware}. Yet gains at the device or array level can shrink substantially once data conversion, peripherals, memory access, synchronization, and other operations are included. There is a still deeper distinction. Improving the physical implementation of an existing operation asks how efficiently a machine can execute an established abstraction. Neuromorphic engineering can ask whether the physics and organization of the machine allow part of that abstraction to be replaced altogether. The opportunity is therefore not only to execute an algorithm with less energy, but also to let the machine's natural dynamics perform the algorithm. A meaningful neuromorphic claim should therefore connect a principle, the physical mechanism that embodies it, the system dynamics it creates, and the capability that emerges.

Much of neuromorphic hardware has progressed bottom-up, from neural primitives toward increasingly complex systems. This Perspective argues for a complementary top-down approach: begin with the capability, identify the organization and dynamics, and choose the substrate. Neuromorphic paradigms should be preserved regardless of the substrate chosen for implementation. This re-framing is important as neuromorphic technologies move toward larger platforms, common software ecosystems, and commercial deployment \cite{kudithipudi2025scale,muir2025commercial}. The abstractions will shape what the field builds and how its progress is judged. A recent call for a revival of neuromorphic engineering emphasizes a return to analog implementations \cite{indiveri2025neuromorphic}. Here we interpret this more broadly as a return to the deeper abstraction that motivated the early field with Mead, Hopfield, and Feynman's \textit{The Physics of Computation}: using the physics, dynamics, and organization of a machine to compute. The following section develops this principle-driven view by identifying principles of nervous-system organization that can lead to better machines.

\section{Neuromorphic Design Principles}

The nervous system produces intelligent behavior by sensing selectively, organizing perception, preserving useful states, and learning from experience. These capabilities are not isolated modules: they arise from the dynamic evolution of interacting physical states. The following principles identify how biological systems turn sensing, memory, learning, and action into useful state evolution, and how artificial systems might be designed to do the same.

\subsection{Sensing what matters}

Engineered sensors are often designed to measure the world as faithfully and uniformly as possible. Biology has a different view: a sensor need not reproduce reality; it may instead be a dynamical system whose internal state changes when the external world becomes relevant. In this sense, sensing is already a form of computation. The nervous system does not record the world uniformly. It selects and transforms signals based on recent history, context, and behavioral need, allocating its limited resources to what is currently informative. The resulting representation is therefore not a faithful copy of the environment, but a compact description suited to perception, learning, and action \cite{simoncelli2001natural,bialek2008efficient}. The intelligence lies in how sensing resources are allocated across signals, channels, locations, and timescales.

\textbf{Adaptation} is fundamental to this selectivity. Natural signals span wide ranges, and their statistics vary continuously. Sensory systems adjust gain, thresholds, and temporal filtering to remain sensitive to informative variations without saturating \cite{fairhall2001efficiency,wark2007sensory}. This is closely related to \textbf{event-based sensing}. When signals are \textbf{sparse}, reporting only significant changes reduces redundant data transfer while preserving temporal information at low latency. In Fig.~\ref{fig:dynamic}, for example, the motion of the bird generates events while much of the scene remains quiet. Event cameras follow this idea by reporting local temporal changes instead of complete image frames \cite{lichtsteiner2008temporal,Gallego2022EventBasedVision}. Biology is more adaptive: recent history and context can change what is considered significant enough to produce a response.

\textbf{Nonlinearity} already plays an important role in sensory organs. The retina compresses and adapts visual signals across a vast range of illumination \cite{Smirnakis1997RetinalAdaptation}. The cochlea does more than simply relay sound; its active mechanics provide frequency selectivity, level-dependent gain, and compression before subsequent neural processing \cite{Dallos2008Cochlear}. These examples challenge the usual engineering tendency to suppress nonlinearity. When appropriately designed, nonlinearity itself can perform useful computation.

Sensing is also \textbf{active}. Where the eyes move changes the visual input, how a surface is touched changes the tactile signal, and sniffing shapes the temporal pattern reaching the olfactory system \cite{schroeder2010active}. The signals available to the nervous system, therefore, depend partly on how they are acquired. An artificial sensor could similarly vary its resolution, bandwidth, sensitivity, or sampling effort with its internal state and task. Sensing then becomes part of the interaction with the environment rather than a fixed front end supplying data.

Biological sensing also relies on \textbf{heterogeneity} and \textbf{population coding}. Elements with different thresholds, gains, selectivity, and timescales respond differently to the same input, while their joint activity carries the representation. In the olfactory bulb, intrinsic diversity among mitral cells can reduce correlated firing and increase population-level information \cite{padmanabhan2010diversity}. \textbf{Redundancy} is not necessarily wasteful. Retinal ganglion cells recorded during natural gaze can retain correlated and redundant responses \cite{karamanlis2025redundant}. Such overlap may provide robustness or preserve features needed downstream. Sparse and redundant representations can therefore both be useful, depending on what the population needs to represent.

\textbf{Stochasticity} can also be useful, but only in the right setting. Most noise is undesirable, yet fluctuations can enhance weak-signal detection in nonlinear systems \cite{douglass1993noise} or provide variability that downstream circuits can use. Its value, therefore, depends on the dynamics in which that variability is embedded.

Finally, sensory processing is distributed over \textbf{multiple timescales}. Fast responses capture immediate changes, adaptation tracks recent signal statistics, and slower feedback adjusts the operating regime over longer periods \cite{wark2007sensory, whitmire2016rapid}. This suggests combining fast signal paths with local state and slower feedback, so that each response reflects both the present input and the recent context.

These mechanisms serve a common purpose: they make sensing selective, stateful, and matched to behavioral demands. A neuromorphic sensor is therefore not a stateless converter. More than providing input to a computation, it determines the state, operating regime, and perturbations from which the subsequent dynamics begin. Once a useful sensory state has been formed, the next question is how it should evolve. This leads from sensing to population dynamics.

\subsection{Perception through collective dynamics}

Perception is the evolution of the collective state under sensory and contextual drive. At any moment, the activity of a neural population can be viewed as a point in a high-dimensional state space. Recurrent interactions determine how that point moves-- whether activity decays, separates into alternatives, circulates, or approaches a stable configuration. The computation is carried out by the \textbf{evolving collective state}. A general description is
\begin{equation}
\dot{\mathbf{x}}
=
\mathbf{F}(\mathbf{x},\mathbf{u},\mathbf{c};\boldsymbol{\theta})
+
\boldsymbol{\eta}(t),
\tag{1}\label{eq1}
\end{equation}
where $\mathbf{x}$ is the collective dynamical state variable that carries information, $\mathbf{u}$ the sensory input that drives the state, $\mathbf{c}$ the context, $\boldsymbol{\theta}$ the slower parameters that learning modifies to shape the dynamics, and $\boldsymbol{\eta}(t)$ the fluctuations. The vector field $\mathbf{F}$ determines how the state evolves. In this view, computation is not only the value of the state at a given moment but also the path the state follows over time. Inputs may initialize or continuously drive trajectories, while context can redirect them through the same recurrent network. Population studies across motor and cognitive tasks show that behavior can be understood from the geometry and evolution of collective activity, rather than from individual neurons \cite{churchland2012neural,mante2013context,vyas2020computation}. In conventional digital systems, the physical trajectory between logical states is largely irrelevant to the computation. A dynamical system makes that trajectory computationally meaningful. The evolution of its state is part of the processing (Fig.~\ref{fig:dynamic}). The transformation can therefore reside both in the represented state and in how that state physically evolves.

Hopfield provided a simple example of this idea. Symmetric recurrent interactions create an energy landscape over the collective state, and the network computes by moving through that landscape toward stable configurations \cite{hopfield1982neural}. Stored memories can be placed as \textbf{attractors}, surrounded by basins that pull nearby states toward them. A partial or corrupted cue, therefore, need not specify a memory address: the recurrent state evolution completes the pattern. Hopfield extended the same principle to continuous, graded neurons and later used it for optimization \cite{hopfield1984neurons, hopfield1985neural}. The same landscape need not be explored only by deterministic descent. In the Boltzmann machine, stochastic transitions allow the state to explore configurations according to their energy, making fluctuations part of the sampling and learning process \cite{ackley1985boltzmann}. Noise can therefore play a constructive role by allowing transitions that deterministic relaxation would not reach. The landscape specifies the available flows; the trajectory is the computation in progress; an attractor and its basins determine which perturbations the system can correct. The deeper idea is not the particular neuron model, but where the computation resides. Memory is distributed across interactions, and recall is a trajectory of the whole network.

This collective view also connects well to modern machine learning. Dense associative memories use stronger nonlinear interactions to create sharper attractor basins with much larger storage capacities, and reveal connections between energy-based associative memories and feedforward networks \cite{krotov2016dense,krotov2021large}. Modern Hopfield networks further connect associative retrieval to the attention operation used in transformers \cite{ramsauer2021hopfield}. Ideas rooted in recurrent network dynamics, therefore, continue to reappear in contemporary artificial intelligence, expressed in different architectures.

Biological networks are often asymmetric, continuously driven, and far from equilibrium. Useful computation can occur through transient amplification, sequences, oscillations, metastable transitions, and rotational flows. Odor representations in the locust antennal lobe, for example, contain informative transient trajectories before reaching more stable activity \cite{mazor2005transient}, while motor-cortical populations during reaching show pronounced rotational dynamics \cite{churchland2012neural}. In context-dependent decisions, \textbf{recurrent population dynamics} can amplify task-relevant evidence while suppressing irrelevant dimensions \cite{mante2013context}. Information may therefore lie in the destination, the path taken, or the timing of transitions. The stable structure itself can take different forms. Continuous attractors preserve an entire manifold of states, allowing continuously valued variables to be maintained while perturbations away from the manifold are corrected. Population activity in \textit{Drosophila} central brain provides an example for heading direction \cite{kim2017ring}. Fixed points, continuous manifolds, metastable states, and transient flows can support different forms of perception, memory, and decision. \textbf{Noise} acquires meaning through this geometry. A perturbation inside a deep attractor basin may be corrected, while that along a weakly stable manifold may cause drift, and near a boundary between alternatives it may trigger a transition \cite{wang2002probabilistic}. Robustness and variability are therefore not opposites: useful dynamics can suppress change along dimensions that should remain stable while retaining sensitivity along dimensions carrying new evidence.

Biology exhibits highly \textbf{distributed computation}, with interactions that are often locally dense but globally sparse. Sparse activity and structured connectivity reduce wiring and communication while preserving population dynamics, allowing global behavior to emerge from selective local and long-range interactions \cite{OlshausenField1996Sparse, MICrONS2025, BullmoreSporns2009Complex}.

For neuromorphic engineering, the design target is therefore the state-space dynamics of the physical network-- its attractors, flows, transition boundaries, timescales, and responses to perturbations. Computation lies in shaping the evolution of the collective state itself. Learning then determines how experience changes that evolution.

\subsection{Learning reshapes the dynamics}

Learning changes the dynamics that govern future behavior. In Eq.~\ref{eq1}, computation changes the state $\mathbf{x}$, whereas learning changes $\boldsymbol{\theta}$. Changes in synaptic strength, excitability, thresholds, gains, delays, or connectivity can deepen an attractor, enlarge its basin, or redirect trajectories through state space. Motor learning, for example, involves changes in both synaptic transmission and intrinsic neuronal properties, while structural plasticity can alter the network topology itself \cite{kida2016motor,xu2009rapid}. Hebb's principle provides the simplest example: correlated activity leaves a lasting change in coupling \cite{hebb1949organization}. From this dynamical perspective, learning is a \textbf{persistent change in collective dynamics} (Fig.~\ref{fig:dynamic}), regardless of whether that change is acquired online or through offline training.

These changes occur over \textbf{multiple timescales}. Short-term synaptic dynamics alter responses over milliseconds to seconds, longer-term plasticity preserves experience, and still slower homeostatic processes regulate the operating regime of the network \cite{tsodyks1997neural,turrigiano1998activity}. Their interaction offers a way to balance plasticity with stability: rapid changes allow adaptation, while slower processes prevent new experience from continually overwriting what has already been learned \cite{fusi2005cascade, benna2016computational, Grossberg1987Competitive}.

Neuromorphic learning, therefore, need not be reduced to programming synaptic weights. Experience may reshape couplings, biases, thresholds, delays, stochasticity, or connectivity, with slower mechanisms stabilizing the resulting dynamics. Not every system needs to learn through biologically local, online plasticity; parameters may be trained offline if they establish the desired dynamical behavior. The same idea can extend into the physics of memory. In learning-in-memory systems, the physical barriers that preserve memory states can also shape how parameters are updated and consolidated \cite{Chen2026EnergyLIM}. Controlling these barriers provides another degree of freedom for engineering the trade-off between retention, adaptation, and the physical cost of updating memory. The broader principle is that the physical dynamics of memory can participate in the learning process itself. Learning is not simply the writing of values into memory. It reshapes the physical constraints through which future computation unfolds.

\subsection{Action closes the loop}

In embodied systems, action is not simply the final output of computation. It changes the relation between the agent and the world, and therefore changes what will be sensed next. Eye movements, whisking, touch, and sniffing all illustrate this principle: behavior helps determine the evidence subsequently available to perception \cite{schroeder2010active, prescott2011active}. Internal state guides action, action reshapes sensory input, and new input updates internal state. Motor-related signals can help active sensing regulate sensory gain, amplifying feedback that is useful for the current action while attenuating predictable or disruptive input \cite{azim2019gain}. Sensory and motor processes, therefore, form a \textbf{continuous dynamical loop} (Fig.~\ref{fig:dynamic}) rather than separate input and output stages. In dynamical terms, action changes the conditions of the next computation: it alters the environment, the sensory drive, and therefore the trajectory available to the system.

The body also participates in this computation. Its geometry, compliance, inertia, and contacts transform both motor commands and sensory signals, sometimes simplifying control or contributing directly to information processing \cite{hauser2012morphological,muller2017morphological}. In this view, embodiment is not simply a processor connected to sensors and actuators. Behavior emerges from the coupled dynamics of the controller, the body, and the environment.

The lesson is to design these elements as a coupled system, coordinating their representations and timescales so that action can direct sensing and new evidence can rapidly alter behavior \cite{bartolozzi2022embodied}. Not every system controls a body, but where action matters, the relevant unit is not always the processor alone; rather, it is the coupled evolution of the agent and the environment. Action does not end computation; it changes what the system will compute next. Part of the computation can be distributed across sensor placement, body mechanics, environmental interaction, and timing of action. Where perception and action are coupled, intelligence belongs not only to the processor, but to the closed-loop dynamical system.

\subsection{Principles of dynamical neuromorphic design}

The preceding discussions suggest a system-level view of neuromorphic engineering in which the central design object is not an artificial neuron, synapse, or device, but the organization and evolution of state across the complete system. This view leads to a practical set of design principles.

\begin{itemize}

\item \textbf{Begin with capability, not components.}
Design should begin by asking what capability the system must produce. Then the states, interactions, dynamics, and representations should be chosen. Starting instead from a neuron model, device, spike representation, or accelerator risks making the implementation the objective.

\item \textbf{Separate principle from substrate, but exploit the substrate.}
A computational principle should not depend on one particular implementation. Adaptation, collective dynamics, or stochastic exploration may be realized in any substrate. Neuromorphic design should therefore abstract the principle while exploiting the medium's native physics.

\item \textbf{Sensing constructs computational state.}
Sensing need not reconstruct the external world uniformly before computation. Adaptation, active sampling, heterogeneity, and population coding can transform physical stimuli directly into states that emphasize what matters for subsequent behavior. Sensing sets the state and operating regime from which later dynamics begin.

\item \textbf{Physical dynamics perform computation in time.}
The evolution of physical state should be shaped so that trajectories perform useful computations. Information may reside in timing, phase, trajectory geometry, or transient evolution. Useful transformations can arise through relaxation, oscillation, transient flow, or stochastic exploration.

\item \textbf{Organize computation collectively.}
Useful computation emerges from many-body interactions rather than from any isolated component. State and computation can be distributed across populations through structured connectivity, sparse activity, and selective event-based communication, without needing dense global coupling or centralized control.

\item \textbf{Memory is encoded in persistent dynamics.}
Memory can reside in persistent activity, slow internal variables, attractor basins, continuous manifolds, synaptic states, or other structures that make the system's future evolution depend on its past. Memory is not just stored information; it sets the trajectories that remain accessible to the system.

\item \textbf{Learning reshapes the dynamical system.}
Learning produces persistent changes in how future states evolve. It may alter couplings, thresholds, gains, time constants, topology, or the geometry of the state space itself. Learning is not just a weight update; it is the process by which experience reorganizes the dynamics that shape future computation.

\item \textbf{Structured variability supports collective computation.}
Biological systems operate with heterogeneity, redundancy, noise, and different timescales. Heterogeneity can diversify, stochasticity can enable exploration, redundancy can improve robustness, and interactions between fast and slow variables can stabilize behavior while preserving adaptability. Variability becomes computationally useful when the surrounding dynamics impose structure on it.

\item \textbf{Close the loop when interaction shapes computation.}
For systems that perceive and act, computation does not end at an output port. Action changes the observed world, which changes what is sensed next. The relevant dynamical system may therefore include the sensor, processor, body, and environment together. Active vision, touch, locomotion, and other forms of embodied sensing exploit this loop to acquire information that passive observation cannot provide.

\end{itemize}

These principles describe ways in which sensing, time, physical dynamics, memory, learning, variability, and interaction can become constitutive parts of computation. Neuromorphic engineering, in this view, is therefore defined by how computation is organized. Its greatest advantage lies in deliberately recruiting the state, dynamics, and physics of a system-- and, where relevant, its interaction with the world-- to perform computational work.

\section{From Principles to Neuromorphic Circuits and Systems}

Many of these principles have already appeared in neuromorphic circuits and systems. Charge can store state; transistor nonlinearities can transform it; feedback can enable adaptation; mismatch can create heterogeneous responses; delays can encode time; recurrent couplings can shape the evolution of many-body states; and physical fluctuations can drive exploration. Neuromorphic circuit design has repeatedly turned such native physical behaviors into useful computation. At the circuit level, this idea produced computational primitives that exploit the exponential current-voltage relationship of subthreshold MOS devices, naturally supporting log-domain and current-mode computation, including tunable first-order Tau-cells, differential-pair integrators, bump-antibump, winner-take-all, and analog correlators \cite{vanSchaik2003TauCell,vanSchaik2010Mihalas, bartolozzi2007synaptic,indiveri2011silicon, delbruck1991bump,lazzaro1989wta}. These small current-mode motifs became building blocks from which larger neuromorphic dynamics could be assembled.

\textbf{Sensing} principles appear most directly in circuits that transform stimuli. In Mead and Mahowald's silicon retina, logarithmic photoreceptors exploited MOS device characteristics to compress a wide illumination range, and a resistive network formed a local spatial average \cite{mead1988silicon}. Adaptive photoreceptors later incorporated slow feedback that shifted the operating point in response to background illumination, combining fast sensing with slower adaptation \cite{mahowald1991adaptive}. Lyon and Mead's silicon cochlea used cascaded second-order analog filters with adaptation to perform continuous frequency and temporal transformations of sound in hardware \cite{lyon1988analog}. The same auditory principles have also been implemented in FPGA, employing cascaded asymmetric resonators, nonlinear feedback, and multi-timescale automatic gain control \cite{xu2018carfac}. Later, cochlear circuits deliberately exploited nonlinear jump resonance and hysteresis arising from the saturating nonlinearity of analog operational transconductance amplifiers \cite{aono2013exploiting}. In event cameras, logarithmic photoreception is followed by temporal differencing and threshold circuits, so a pixel communicates only when local intensity changes sufficiently \cite{lichtsteiner2008temporal}. Adaptation, nonlinear filtering, multiple timescales, and selective communication, therefore, begin within the sensory front end. Heterogeneity and stochasticity have also been turned into resources. The Trainable Analog Block exploits transistor mismatch to generate a heterogeneous population of nonlinear responses and high-dimensional projections, while trainable output weights learn to combine them by using their diversity \cite{thakur2016tab,thakur2018tab}. Noise has been used physically in stochastic-resonance circuits, probabilistic neural computation, and stochastic electronics, reducing complex operations to simple logic gates \cite{querlioz2013stochastic,chien2018sampling,hamilton2014stochastic}. These circuits have supported hardware implementations of Bayesian estimation, inference, and visual saliency \cite{thakur2016bayesian,thakur2017saliency}. These examples show how the sensing principle appears before data become conventional representations.

The same logic extends from sensing front-ends to \textbf{recurrent networks}, in which the circuit trajectory itself performs the computation. Hopfield networks were implemented with recurrent feedback circuits that guide the collective state to relax toward configurations minimizing optimization objectives \cite{hopfield1982neural,hopfield1985neural}. The Boltzmann machine dynamics have also been mapped onto dedicated digital hardware for accelerated sampling
\cite{ackley1985boltzmann, LyChow2010RBMFPGA, Patel2022HardwareRBM}. Modern hardware such as p-bits provides fluctuating binary states \cite{camsari2017pbits}, memristor Hopfield networks exploit intrinsic device noise \cite{cai2020intrinsic}, and neuromorphic Ising machines implemented on platforms such as SpiNNaker and FPGAs use coupled asynchronous dynamics with controlled fluctuations to search optimization landscapes \cite{chen2025neurosa,ahsan2026higherorder}. Mixed-signal systems such as ROLLS, Neurogrid, DYNAPs, and BrainScaleS implement evolving analog neural and synaptic states, coupled with digital events \cite{qiao2015rolls, Benjamin2014Neurogrid, Moradi2018DYNAPs, Pehle2022BrainScaleS2}. Fully digital processors such as SpiNNaker, Loihi, and FPGA implementations of large spiking systems show that the recurrent and temporal organization can be realized even without analog neurons \cite{Furber2014SpiNNaker, Davies2018Loihi, wang2018fpga}.

Even communication can become computational. Address-event representation (AER) established a practical means of communicating sparse neural activity \cite{boahen2000aer}. Asynchronous neuromorphic hardware employs event-driven, quasi-delay-insensitive circuits and local handshakes and has built scalable digital systems such as TrueNorth \cite{imam2012qdi,li2025neuroscale,merolla2014truenorth}. Processing-in-Interconnect treats delays in routers and switches as computational primitives \cite{Srivatsav2025Pi2}. In such systems, communication is not merely an overhead; its timing and dynamics can shape the computational transformation.

\textbf{Learning} changes the future dynamics of the circuit. Floating-gate MOS transistors demonstrated compact, non-volatile analog synaptic state  \cite{diorio1996synapse}. Hebbian and spike-timing-dependent plasticity (STDP) circuits subsequently allowed neural activity itself to modify synaptic state, implemented in systems such as BrainScaleS and stochastic neuromorphic hardware \cite{hebb1949organization,indiveri2006vlsi,friedmann2017hybrid, Pehle2022BrainScaleS2, wang2016stochasticSTDP}. FPGAs have been used to train Boltzmann networks for classification \cite{niazi2024boltzmann}.

Finally, these principles can be embedded in closed-loop \textbf{action}. Sensor-processor co-design integrates sensing, dynamics, learning, and action into a single physical feedback loop rather than as independent stages \cite{milde2017obstacle,glatz2019adaptive}.

Altogether, these examples span subthreshold current-mode circuits, mixed-signal integrated circuits, asynchronous digital processors, FPGAs, and emerging stochastic devices. Across substrates, the central objective is the same: to identify the physical properties that can contribute to computation and deliberately organize them for useful function. Any claimed advantage must therefore be assessed at the system boundary \cite{yik2025neurobench}. When the intrinsic dynamics of the substrate are exploited in this way, the hardware is no longer merely a platform on which an algorithm runs; its physical evolution contributes directly to the computation.

\section{Final Outlook}

Neuromorphic engineering is entering a consequential phase. Event-driven sensors, large-scale processors, mixed-signal systems, emerging memory devices, and increasingly mature software are bringing the field closer to large-scale practical deployment. With that maturity comes a risk: the field may gradually become defined by the technologies it has learned to build. If neuromorphic comes to mean only spikes, artificial synapses, or efficient hardware for neural-network workloads, it may succeed as a branch of AI hardware while leaving its deeper scientific opportunity unexplored. The question is no longer simply how far these technologies can scale, but toward what idea of computation they are being scaled.

Reviving neuromorphic engineering does not necessarily require returning to an earlier circuit style, a particular neuron model, or a preferred substrate. It requires recovering a broader abstraction. Biology is not a catalog of components to reproduce, but the evidence that efficient computation can arise from forms of organization very different from those on which conventional computers were built. A biological mechanism becomes useful to engineering when one can identify the computational principle it embodies, understand why that principle matters, and determine how to realize it in another physical system.

In this view, the theme of neuromorphic design is the organization of dynamics. This Perspective returns to the fundamental thread near the field's beginnings, reflected in Mead, Hopfield, and Feynman's \textit{The Physics of Computation}, in which computation is inseparable from the physical laws, state-space dynamics, and constraints of the system performing it.

This framing gives neuromorphic engineering a scientific scope broader than just neural acceleration. It invites neuroscience to reveal principles, physics to supply new computational degrees of freedom, and engineering to turn them into systems whose capabilities emerge from their organization and interaction with the world. The enduring promise of neuromorphic engineering is not to reproduce the nervous system, but to let what we learn from it expand our idea of what a computing machine can sense, remember, learn, and become.

\section*{Acknowledgments}
AI tools were used to restructure the sentences and paragraphs and to correct for typographical/grammatical errors.

\section*{Statements and Declarations}

\subsection*{Competing Interests}
On behalf of all authors, the corresponding author states that there is no conflict of interest.

\bibliography{sn-bibliography}

@article{LopezMunoz2006Neuron,
  author       = {Francisco López-Muñoz and Jesús Boya and Cecilio Alamo},
  title        = {Neuron theory, the cornerstone of neuroscience, on the centenary of the Nobel Prize award to Santiago Ramón y Cajal},
  journal      = {Brain Research Bulletin},
  year         = {2006},
  volume       = {70},
  number       = {4--6},
  pages        = {391--405},
  doi          = {10.1016/j.brainresbull.2006.07.010}
}

@article{HodgkinHuxley1952,
  author       = {Hodgkin, A. L. and Huxley, A. F.},
  title        = {A quantitative description of membrane current and its application to conduction and excitation in nerve},
  journal      = {The Journal of Physiology},
  year         = {1952},
  volume       = {117},
  number       = {4},
  pages        = {500--544},
  doi          = {10.1113/jphysiol.1952.sp004764}
}

@article{Roland2023HowFar,
  author       = {Roland, Per E.},
  title        = {How far neuroscience is from understanding brains},
  journal      = {Frontiers in Systems Neuroscience},
  year         = {2023},
  volume       = {17},
  pages        = {1147896},
  doi          = {10.3389/fnsys.2023.1147896}
}

@article{DelbruckLiuMead2026,
  author       = {Delbruck, Tobi and Liu, Shih-Chii},
  title        = {A Conversation With Carver Mead About the Caltech “Neuromorphic Era” and What Came After [Interview]},
  journal      = {IEEE Circuits and Systems Magazine},
  year         = {2026},
  volume       = {26},
  number       = {3},
  pages        = {3--12},
  doi          = {10.1109/MCAS.2026.3695028}
}

@article{VittozFellrath1977,
  author       = {Vittoz, E. and Fellrath, J.},
  title        = {CMOS analog integrated circuits based on weak inversion operations},
  journal      = {IEEE Journal of Solid-State Circuits},
  year         = {1977},
  volume       = {12},
  number       = {3},
  pages        = {224--231},
  doi          = {10.1109/JSSC.1977.1050882}
}

@book{feynman2018feynman,
  author       = {Feynman, Richard P.},
  title        = {Feynman Lectures on Computation},
  series       = {Frontiers in Physics},
  publisher    = {CRC Press},
  year         = {2018},
  isbn         = {9780429968990},
  doi          = {10.1201/9780429500442}
}

@misc{mead2022oralhistory,
  author       = {Mead, Carver A.},
  title        = {Carver Mead (BS '56, PhD '60), Electrical Engineer},
  year         = {2022},
  month        = {jun},
  howpublished = {Caltech Heritage Project},
  note         = {Oral history interview by David Zierler; interviews conducted June--August 2020},
  url          = {https://heritageproject.caltech.edu/interviews-updates/carver-mead},
  urldate      = {2026-08-24}
}

@article{hopfield1982neural,
  author       = {J J Hopfield},
  title        = {Neural networks and physical systems with emergent collective computational abilities.},
  journal      = {Proceedings of the National Academy of Sciences},
  year         = {1982},
  volume       = {79},
  number       = {8},
  pages        = {2554--2558},
  doi          = {10.1073/pnas.79.8.2554}
}

@article{mead2023mahowald,
  author       = {Mead, Carver},
  title        = {Neuromorphic Engineering: In Memory of Misha Mahowald},
  journal      = {Neural Computation},
  year         = {2023},
  volume       = {35},
  number       = {3},
  pages        = {343--383},
  doi          = {10.1162/neco_a_01553}
}

@article{mead1988silicon,
  author       = {Carver A. Mead and M.A. Mahowald},
  title        = {A silicon model of early visual processing},
  journal      = {Neural Networks},
  year         = {1988},
  volume       = {1},
  number       = {1},
  pages        = {91--97},
  doi          = {10.1016/0893-6080(88)90024-X}
}

@phdthesis{mahowald1992vlsi,
  author       = {Mahowald, Michelle A. (Misha)},
  title        = {VLSI analogs of neuronal visual processing: a synthesis of form and function},
  school       = {California Institute of Technology},
  year         = {1992},
  doi          = {10.7907/4bdw-fg34},
  publisher    = {California Institute of Technology},
  month        = {may}
}

@book{mahowald1994stereoscopic,
  author       = {Mahowald, M.},
  title        = {An Analog VLSI System for Stereoscopic Vision},
  series       = {The Springer International Series in Engineering and Computer Science},
  publisher    = {Springer US},
  year         = {1994},
  isbn         = {9780792394440},
  doi          = {10.1007/978-1-4615-2724-4},
  volume       = {265}
}

@article{mead1990neuromorphic,
  author       = {Mead, C.},
  title        = {Neuromorphic electronic systems},
  journal      = {Proceedings of the IEEE},
  year         = {1990},
  volume       = {78},
  number       = {10},
  pages        = {1629--1636},
  doi          = {10.1109/5.58356}
}

@book{mead1989analog,
  author       = {Mead, Carver A.},
  title        = {Analog VLSI and Neural Systems},
  publisher    = {Addison-Wesley},
  address      = {Reading, MA},
  year         = {1989},
  isbn         = {978-0-201-05992-2},
  url          = {https://authors.library.caltech.edu/records/0pwma-g7y55}
}

@article{Amari1972LearningPatterns,
  author       = {Amari, S.-I.},
  title        = {Learning Patterns and Pattern Sequences by Self-Organizing Nets of Threshold Elements},
  journal      = {IEEE Transactions on Computers},
  year         = {1972},
  volume       = {C-21},
  number       = {11},
  pages        = {1197--1206},
  doi          = {10.1109/T-C.1972.223477}
}

@article{Nakano1972Associatron,
  author       = {Nakano, Kaoru},
  title        = {Associatron-A Model of Associative Memory},
  journal      = {IEEE Transactions on Systems, Man, and Cybernetics},
  year         = {1972},
  volume       = {SMC-2},
  number       = {3},
  pages        = {380--388},
  doi          = {10.1109/TSMC.1972.4309133}
}

@article{Little1974PersistentStates,
  author       = {W.A. Little},
  title        = {The existence of persistent states in the brain},
  journal      = {Mathematical Biosciences},
  year         = {1974},
  volume       = {19},
  number       = {1},
  pages        = {101--120},
  doi          = {10.1016/0025-5564(74)90031-5}
}

@article{Amari1967AdaptivePattern,
  author       = {Amari, Shunichi},
  title        = {A Theory of Adaptive Pattern Classifiers},
  journal      = {IEEE Transactions on Electronic Computers},
  year         = {1967},
  volume       = {EC-16},
  number       = {3},
  pages        = {299--307},
  doi          = {10.1109/PGEC.1967.264666}
}

@article{Fukushima1969VisualFeature,
  author       = {Fukushima, Kunihiko},
  title        = {Visual Feature Extraction by a Multilayered Network of Analog Threshold Elements},
  journal      = {IEEE Transactions on Systems Science and Cybernetics},
  year         = {1969},
  volume       = {5},
  number       = {4},
  pages        = {322--333},
  doi          = {10.1109/TSSC.1969.300225}
}

@article{Fukushima1980Neocognitron,
  author       = {Fukushima, Kunihiko},
  title        = {Neocognitron: A self-organizing neural network model for a mechanism of pattern recognition unaffected by shift in position},
  journal      = {Biological Cybernetics},
  year         = {1980},
  volume       = {36},
  number       = {4},
  pages        = {193--202},
  doi          = {10.1007/BF00344251}
}

@article{Benjamin2014Neurogrid,
  author       = {Benjamin, Ben Varkey and Gao, Peiran and McQuinn, Emmett and Choudhary, Swadesh and Chandrasekaran, Anand R. and Bussat, Jean-Marie and Alvarez-Icaza, Rodrigo and Arthur, John V. and Merolla, Paul A. and Boahen, Kwabena},
  title        = {Neurogrid: A Mixed-Analog-Digital Multichip System for Large-Scale Neural Simulations},
  journal      = {Proceedings of the IEEE},
  year         = {2014},
  volume       = {102},
  number       = {5},
  pages        = {699--716},
  doi          = {10.1109/JPROC.2014.2313565}
}

@article{Pehle2022BrainScaleS2,
  author       = {Pehle, Christian and Billaudelle, Sebastian and Cramer, Benjamin and Kaiser, Jakob and Schreiber, Korbinian and Stradmann, Yannik and Weis, Johannes and Leibfried, Aron and Müller, Eric and Schemmel, Johannes},
  title        = {The BrainScaleS-2 Accelerated Neuromorphic System With Hybrid Plasticity},
  journal      = {Frontiers in Neuroscience},
  year         = {2022},
  volume       = {16},
  pages        = {795876},
  doi          = {10.3389/fnins.2022.795876}
}

@article{Furber2014SpiNNaker,
  author       = {Furber, Steve B. and Galluppi, Francesco and Temple, Steve and Plana, Luis A.},
  title        = {The SpiNNaker Project},
  journal      = {Proceedings of the IEEE},
  year         = {2014},
  volume       = {102},
  number       = {5},
  pages        = {652--665},
  doi          = {10.1109/JPROC.2014.2304638}
}

@article{Davies2018Loihi,
  author       = {Davies, Mike and Srinivasa, Narayan and Lin, Tsung-Han and Chinya, Gautham and Cao, Yongqiang and Choday, Sri Harsha and Dimou, Georgios and Joshi, Prasad and Imam, Nabil and Jain, Shweta and Liao, Yuyun and Lin, Chit-Kwan and Lines, Andrew and Liu, Ruokun and Mathaikutty, Deepak and McCoy, Steven and Paul, Arnab and Tse, Jonathan and Venkataramanan, Guruguhanathan and Weng, Yi-Hsin and Wild, Andreas and Yang, Yoonseok and Wang, Hong},
  title        = {Loihi: A Neuromorphic Manycore Processor with On-Chip Learning},
  journal      = {IEEE Micro},
  year         = {2018},
  volume       = {38},
  number       = {1},
  pages        = {82--99},
  doi          = {10.1109/MM.2018.112130359}
}

@misc{Srivatsav2025Pi2,
  author       = {Srivatsav R, Madhuvanthi and Bhattacharyya, Chiranjib and Chakrabartty, Shantanu and Thakur, Chetan Singh},
  title        = {When Routers, Switches and Interconnects Compute: A processing-in-interconnect Paradigm for Scalable Neuromorphic AI},
  year         = {2025},
  eprint       = {2508.19548},
  archiveprefix = {arXiv},
  primaryclass = {cs.ET},
  doi          = {10.48550/arXiv.2508.19548}
}

@article{MICrONS2025,
  author       = {{{The MICrONS Consortium}}},
  title        = {Functional connectomics spanning multiple areas of mouse visual cortex},
  journal      = {Nature},
  year         = {2025},
  volume       = {640},
  number       = {8058},
  pages        = {435--447},
  doi          = {10.1038/s41586-025-08790-w}
}

@article{christensen2022roadmap,
  author       = {Christensen, Dennis V and Dittmann, Regina and Linares-Barranco, Bernabe and Sebastian, Abu and Le Gallo, Manuel and Redaelli, Andrea and Slesazeck, Stefan and Mikolajick, Thomas and Spiga, Sabina and Menzel, Stephan and others},
  title        = {2022 roadmap on neuromorphic computing and engineering},
  journal      = {Neuromorphic Computing and Engineering},
  year         = {2022},
  volume       = {2},
  number       = {2},
  pages        = {022501},
  doi          = {10.1088/2634-4386/ac4a83}
}

@article{kudithipudi2025scale,
  author       = {Kudithipudi, Dhireesha and Schuman, Catherine and Vineyard, Craig M and Pandit, Tej and Merkel, Cory and Kubendran, Rajkumar and Aimone, James B and Orchard, Garrick and Mayr, Christian and Benosman, Ryad and others},
  title        = {Neuromorphic computing at scale},
  journal      = {Nature},
  year         = {2025},
  volume       = {637},
  number       = {8047},
  pages        = {801--812},
  doi          = {10.1038/s41586-024-08253-8}
}

@article{schuman2022opportunities,
  author       = {Schuman, Catherine D and Kulkarni, Shruti R and Parsa, Maryam and Mitchell, J Parker and Date, Prasanna and Kay, Bill},
  title        = {Opportunities for neuromorphic computing algorithms and applications},
  journal      = {Nature Computational Science},
  year         = {2022},
  volume       = {2},
  number       = {1},
  pages        = {10--19},
  doi          = {10.1038/s43588-021-00184-y}
}

@article{sebastian2020memory,
  author       = {Sebastian, Abu and Le Gallo, Manuel and Khaddam-Aljameh, Riduan and Eleftheriou, Evangelos},
  title        = {Memory devices and applications for in-memory computing},
  journal      = {Nature Nanotechnology},
  year         = {2020},
  volume       = {15},
  number       = {7},
  pages        = {529--544},
  doi          = {10.1038/s41565-020-0655-z}
}

@article{aguirre2024hardware,
  author       = {Aguirre, Fernando and Sebastian, Abu and Le Gallo, Manuel and Song, Wenhao and Wang, Tong and Yang, J Joshua and Lu, Wei and Chang, Meng-Fan and Ielmini, Daniele and Yang, Yuchao and others},
  title        = {Hardware implementation of memristor-based artificial neural networks},
  journal      = {Nature Communications},
  year         = {2024},
  volume       = {15},
  number       = {1},
  pages        = {1974},
  doi          = {10.1038/s41467-024-45670-9}
}

@article{muir2025commercial,
  author       = {Muir, Dylan Richard and Sheik, Sadique},
  title        = {The road to commercial success for neuromorphic technologies},
  journal      = {Nature Communications},
  year         = {2025},
  volume       = {16},
  number       = {1},
  pages        = {3586},
  doi          = {10.1038/s41467-025-57352-1}
}

@article{indiveri2025neuromorphic,
  author       = {Indiveri, Giacomo},
  title        = {Neuromorphic is dead. Long live neuromorphic.},
  journal      = {Neuron},
  year         = {2025},
  volume       = {113},
  number       = {20},
  pages        = {3311--3314},
  doi          = {10.1016/j.neuron.2025.09.020}
}

@article{simoncelli2001natural,
  author       = {Simoncelli, Eero P and Olshausen, Bruno A},
  title        = {Natural image statistics and neural representation},
  journal      = {Annual Review of Neuroscience},
  year         = {2001},
  volume       = {24},
  number       = {1},
  pages        = {1193--1216},
  doi          = {10.1146/annurev.neuro.24.1.1193}
}

@inproceedings{bialek2008efficient,
  author       = {Bialek, William and De Ruyter Van Steveninck, Rob R. and Tishby, Naftali},
  title        = {Efficient representation as a design principle for neural coding and computation},
  booktitle    = {2006 IEEE International Symposium on Information Theory},
  year         = {2006},
  pages        = {659--663},
  doi          = {10.1109/ISIT.2006.261867}
}

@article{fairhall2001efficiency,
  author       = {Fairhall, Adrienne L and Lewen, Geoffrey D and Bialek, William and de Ruyter van Steveninck, Robert R},
  title        = {Efficiency and ambiguity in an adaptive neural code},
  journal      = {Nature},
  year         = {2001},
  volume       = {412},
  number       = {6849},
  pages        = {787--792},
  doi          = {10.1038/35090500}
}

@article{wark2007sensory,
  author       = {Barry Wark and Brian Nils Lundstrom and Adrienne Fairhall},
  title        = {Sensory adaptation},
  journal      = {Current Opinion in Neurobiology},
  year         = {2007},
  volume       = {17},
  number       = {4},
  pages        = {423--429},
  doi          = {10.1016/j.conb.2007.07.001}
}

@article{lichtsteiner2008temporal,
  author       = {Lichtsteiner, Patrick and Posch, Christoph and Delbruck, Tobi},
  title        = {A 128$\times$ 128 120 dB 15 $\mu$s Latency Asynchronous Temporal Contrast Vision Sensor},
  journal      = {IEEE Journal of Solid-State Circuits},
  year         = {2008},
  volume       = {43},
  number       = {2},
  pages        = {566--576},
  doi          = {10.1109/JSSC.2007.914337}
}

@article{Gallego2022EventBasedVision,
  author  = {Gallego, Guillermo and
             Delbruck, Tobi and
             Orchard, Garrick and
             Bartolozzi, Chiara and
             Taba, Brian and
             Censi, Andrea and
             Leutenegger, Stefan and
             Davison, Andrew J. and
             Conradt, Jorg and
             Daniilidis, Kostas and
             Scaramuzza, Davide},
  title   = {Event-Based Vision: A Survey},
  journal = {IEEE Transactions on Pattern Analysis and Machine Intelligence},
  year    = {2022},
  volume  = {44},
  number  = {1},
  pages   = {154--180},
  doi     = {10.1109/TPAMI.2020.3008413},
  url     = {https://doi.org/10.1109/TPAMI.2020.3008413}
}

@article{Smirnakis1997RetinalAdaptation,
  author  = {Smirnakis, Stelios M. and
             Berry, Michael J. and
             Warland, David K. and
             Bialek, William and
             Meister, Markus},
  title   = {Adaptation of Retinal Processing to Image Contrast and Spatial Scale},
  journal = {Nature},
  year    = {1997},
  volume  = {386},
  pages   = {69--73},
  doi     = {10.1038/386069a0}
}

@article{Dallos2008Cochlear,
  author       = {Peter Dallos},
  title        = {Cochlear amplification, outer hair cells and prestin},
  journal      = {Current Opinion in Neurobiology},
  year         = {2008},
  volume       = {18},
  number       = {4},
  pages        = {370--376},
  doi          = {10.1016/j.conb.2008.08.016}
}

@article{schroeder2010active,
  author       = {Charles E Schroeder and Donald A Wilson and Thomas Radman and Helen Scharfman and Peter Lakatos},
  title        = {Dynamics of Active Sensing and perceptual selection},
  journal      = {Current Opinion in Neurobiology},
  year         = {2010},
  volume       = {20},
  number       = {2},
  pages        = {172--176},
  doi          = {10.1016/j.conb.2010.02.010}
}

@article{padmanabhan2010diversity,
  author       = {Padmanabhan, Krishnan and Urban, Nathaniel N},
  title        = {Intrinsic biophysical diversity decorrelates neuronal firing while increasing information content},
  journal      = {Nature Neuroscience},
  year         = {2010},
  volume       = {13},
  number       = {10},
  pages        = {1276--1282},
  doi          = {10.1038/nn.2630}
}

@article{karamanlis2025redundant,
  author       = {Karamanlis, Dimokratis and Khani, Mohammad H and Schreyer, Helene M and Zapp, S{\"o}ren J and Mietsch, Matthias and Gollisch, Tim},
  title        = {Nonlinear receptive fields evoke redundant retinal coding of natural scenes},
  journal      = {Nature},
  year         = {2025},
  volume       = {637},
  number       = {8045},
  pages        = {394--401},
  doi          = {10.1038/s41586-024-08212-3}
}

@article{douglass1993noise,
  author       = {Douglass, John K and Wilkens, Lon and Pantazelou, Eleni and Moss, Frank},
  title        = {Noise enhancement of information transfer in crayfish mechanoreceptors by stochastic resonance},
  journal      = {Nature},
  year         = {1993},
  volume       = {365},
  number       = {6444},
  pages        = {337--340},
  doi          = {10.1038/365337a0}
}

@article{whitmire2016rapid,
  author       = {Whitmire, Clarissa J and Stanley, Garrett B},
  title        = {Rapid sensory adaptation redux: a circuit perspective},
  journal      = {Neuron},
  year         = {2016},
  volume       = {92},
  number       = {2},
  pages        = {298--315},
  doi          = {10.1016/j.neuron.2016.09.046}
}

@article{churchland2012neural,
  author       = {Churchland, Mark M and Cunningham, John P and Kaufman, Matthew T and Foster, Justin D and Nuyujukian, Paul and Ryu, Stephen I and Shenoy, Krishna V},
  title        = {Neural population dynamics during reaching},
  journal      = {Nature},
  year         = {2012},
  volume       = {487},
  number       = {7405},
  pages        = {51--56},
  doi          = {10.1038/nature11129}
}

@article{mante2013context,
  author       = {Mante, Valerio and Sussillo, David and Shenoy, Krishna V and Newsome, William T},
  title        = {Context-dependent computation by recurrent dynamics in prefrontal cortex},
  journal      = {Nature},
  year         = {2013},
  volume       = {503},
  number       = {7474},
  pages        = {78--84},
  doi          = {10.1038/nature12742}
}

@article{vyas2020computation,
  author       = {Vyas, Saurabh and Golub, Matthew D and Sussillo, David and Shenoy, Krishna V},
  title        = {Computation through neural population dynamics},
  journal      = {Annual Review of Neuroscience},
  year         = {2020},
  volume       = {43},
  number       = {1},
  pages        = {249--275},
  doi          = {10.1146/annurev-neuro-092619-094115}
}

@article{hopfield1984neurons,
  author       = {J J Hopfield},
  title        = {Neurons with graded response have collective computational properties like those of two-state neurons.},
  journal      = {Proceedings of the National Academy of Sciences},
  year         = {1984},
  volume       = {81},
  number       = {10},
  pages        = {3088--3092},
  doi          = {10.1073/pnas.81.10.3088},
  eprint       = {https://www.pnas.org/doi/pdf/10.1073/pnas.81.10.3088}
}

@article{hopfield1985neural,
  author       = {Hopfield, John J and Tank, David W},
  title        = {“Neural” computation of decisions in optimization problems},
  journal      = {Biological Cybernetics},
  year         = {1985},
  volume       = {52},
  number       = {3},
  pages        = {141--152},
  doi          = {10.1007/BF00339943}
}

@article{ackley1985boltzmann,
  author       = {David H. Ackley and Geoffrey E. Hinton and Terrence J. Sejnowski},
  title        = {A learning algorithm for boltzmann machines},
  journal      = {Cognitive Science},
  year         = {1985},
  volume       = {9},
  number       = {1},
  pages        = {147--169},
  doi          = {10.1016/S0364-0213(85)80012-4}
}

@inproceedings{krotov2016dense,
  author       = {Krotov, Dmitry and Hopfield, John J.},
  title        = {Dense Associative Memory for Pattern Recognition},
  booktitle    = {Advances in Neural Information Processing Systems},
  year         = {2016},
  volume       = {29},
  pages        = {1172--1180},
  editor       = {D. Lee and M. Sugiyama and U. Luxburg and I. Guyon and R. Garnett},
  publisher    = {Curran Associates, Inc.},
  url          = {https://proceedings.neurips.cc/paper\_files/paper/2016/file/eaae339c4d89fc102edd9dbdb6a28915-Paper.pdf}
}

@inproceedings{krotov2021large,
  author       = {Dmitry Krotov and John J. Hopfield},
  title        = {Large Associative Memory Problem in Neurobiology and Machine Learning},
  booktitle    = {International Conference on Learning Representations},
  year         = {2021},
  url          = {https://openreview.net/forum?id=X4y\_10OX-hX}
}

@inproceedings{ramsauer2021hopfield,
  author       = {Hubert Ramsauer and Bernhard Sch{\"a}fl and Johannes Lehner and Philipp Seidl and Michael Widrich and Lukas Gruber and Markus Holzleitner and Thomas Adler and David Kreil and Michael K Kopp and G{\"u}nter Klambauer and Johannes Brandstetter and Sepp Hochreiter},
  title        = {Hopfield Networks is All You Need},
  booktitle    = {International Conference on Learning Representations},
  year         = {2021},
  url          = {https://openreview.net/forum?id=tL89RnzIiCd}
}

@article{mazor2005transient,
  author       = {Mazor, Ofer and Laurent, Gilles},
  title        = {Transient dynamics versus fixed points in odor representations by locust antennal lobe projection neurons},
  journal      = {Neuron},
  year         = {2005},
  volume       = {48},
  number       = {4},
  pages        = {661--673},
  doi          = {10.1016/j.neuron.2005.09.032}
}

@article{kim2017ring,
  author       = {Sung Soo Kim and Hervé Rouault and Shaul Druckmann and Vivek Jayaraman},
  title        = {Ring attractor dynamics in the Drosophila central brain},
  journal      = {Science},
  year         = {2017},
  volume       = {356},
  number       = {6340},
  pages        = {849--853},
  doi          = {10.1126/science.aal4835}
}

@article{wang2002probabilistic,
  author       = {Wang, Xiao-Jing},
  title        = {Probabilistic decision making by slow reverberation in cortical circuits},
  journal      = {Neuron},
  year         = {2002},
  volume       = {36},
  number       = {5},
  pages        = {955--968},
  doi          = {10.1016/S0896-6273(02)01092-9}
}

@article{OlshausenField1996Sparse,
  author       = {Olshausen, Bruno A and Field, David J},
  title        = {Emergence of simple-cell receptive field properties by learning a sparse code for natural images},
  journal      = {Nature},
  year         = {1996},
  volume       = {381},
  number       = {6583},
  pages        = {607--609},
  doi          = {10.1038/381607a0}
}

@article{BullmoreSporns2009Complex,
  author  = {Bullmore, Ed and Sporns, Olaf},
  title   = {Complex Brain Networks: Graph Theoretical Analysis of Structural and Functional Systems},
  journal = {Nature Reviews Neuroscience},
  year    = {2009},
  volume  = {10},
  number  = {3},
  pages   = {186--198},
  doi     = {10.1038/nrn2575},
  url     = {https://doi.org/10.1038/nrn2575}
}

@article{kida2016motor,
  author       = {Kida, Hiroyuki and Tsuda, Yasumasa and Ito, Nana and Yamamoto, Yui and Owada, Yuji and Kamiya, Yoshinori and Mitsushima, Dai},
  title        = {Motor Training Promotes Both Synaptic and Intrinsic Plasticity of Layer II/III Pyramidal Neurons in the Primary Motor Cortex},
  journal      = {Cerebral Cortex},
  year         = {2016},
  volume       = {26},
  number       = {8},
  pages        = {3494--3507},
  month        = {08},
  doi          = {10.1093/cercor/bhw134}
}

@article{xu2009rapid,
  author       = {Xu, Tonghui and Yu, Xinzhu and Perlik, Andrew J and Tobin, Willie F and Zweig, Jonathan A and Tennant, Kelly and Jones, Theresa and Zuo, Yi},
  title        = {Rapid formation and selective stabilization of synapses for enduring motor memories},
  journal      = {Nature},
  year         = {2009},
  volume       = {462},
  number       = {7275},
  pages        = {915--919},
  doi          = {10.1038/nature08389}
}

@book{hebb1949organization,
  author       = {Hebb, Donald O.},
  title        = {The Organization of Behavior: A Neuropsychological Theory},
  publisher    = {John Wiley \& Sons},
  address      = {New York},
  year         = {1949},
  url          = {https://hdl.handle.net/11858/00-001M-0000-002D-FFA5-4}
}

@article{tsodyks1997neural,
  author       = {Misha V. Tsodyks and Henry Markram},
  title        = {The neural code between neocortical pyramidal neurons depends on neurotransmitter release probability},
  journal      = {Proceedings of the National Academy of Sciences},
  year         = {1997},
  volume       = {94},
  number       = {2},
  pages        = {719--723},
  doi          = {10.1073/pnas.94.2.719},
  eprint       = {https://www.pnas.org/doi/pdf/10.1073/pnas.94.2.719}
}

@article{turrigiano1998activity,
  author       = {Turrigiano, Gina G and Leslie, Kenneth R and Desai, Niraj S and Rutherford, Lana C and Nelson, Sacha B},
  title        = {Activity-dependent scaling of quantal amplitude in neocortical neurons},
  journal      = {Nature},
  year         = {1998},
  volume       = {391},
  number       = {6670},
  pages        = {892--896},
  doi          = {10.1038/36103}
}

@article{fusi2005cascade,
  author       = {Fusi, Stefano and Drew, Patrick J and Abbott, Larry F},
  title        = {Cascade models of synaptically stored memories},
  journal      = {Neuron},
  year         = {2005},
  volume       = {45},
  number       = {4},
  pages        = {599--611},
  doi          = {10.1016/j.neuron.2005.02.001}
}

@article{benna2016computational,
  author       = {Benna, Marcus K and Fusi, Stefano},
  title        = {Computational principles of synaptic memory consolidation},
  journal      = {Nature Neuroscience},
  year         = {2016},
  volume       = {19},
  number       = {12},
  pages        = {1697--1706},
  doi          = {10.1038/nn.4401}
}

@article{Grossberg1987Competitive,
  author  = {Grossberg, Stephen},
  title   = {Competitive Learning: From Interactive Activation to Adaptive Resonance},
  journal = {Cognitive Science},
  year    = {1987},
  volume  = {11},
  number  = {1},
  pages   = {23--63},
  doi     = {10.1016/S0364-0213(87)80025-3}
}

@article{Chen2026EnergyLIM,
  author       = {Chen, Zihao and Ahsan, Faiek and Chakrabartty, Shantanu and Leugering, Johannes and Cauwenberghs, Gert},
  title        = {Estimation of energy-dissipation lower bounds for neuromorphic learning in memory},
  journal      = {Physical Review E},
  year         = {2026},
  volume       = {113},
  number       = {3},
  pages        = {035311},
  month        = {Mar},
  doi          = {10.1103/497f-nt8h}
}

@article{prescott2011active,
  author       = {Prescott, Tony J and Diamond, Mathew E and Wing, Alan M},
  title        = {Active touch sensing},
  journal      = {Philosophical Transactions of the Royal Society B: Biological Sciences},
  year         = {2011},
  volume       = {366},
  number       = {1581},
  pages        = {2989--2995},
  doi          = {10.1098/rstb.2011.0167}
}

@article{azim2019gain,
  author       = {Eiman Azim and Kazuhiko Seki},
  title        = {Gain control in the sensorimotor system},
  journal      = {Current Opinion in Physiology},
  year         = {2019},
  volume       = {8},
  pages        = {177--187},
  doi          = {10.1016/j.cophys.2019.03.005}
}

@article{hauser2012morphological,
  author       = {Hauser, Helmut and Ijspeert, Auke J and F{\"u}chslin, Rudolf M and Pfeifer, Rolf and Maass, Wolfgang},
  title        = {Towards a theoretical foundation for morphological computation with compliant bodies},
  journal      = {Biological Cybernetics},
  year         = {2011},
  volume       = {105},
  number       = {5--6},
  pages        = {355--370},
  doi          = {10.1007/s00422-012-0471-0}
}

@article{muller2017morphological,
  author       = {Müller, Vincent C. and Hoffmann, Matej},
  title        = {What Is Morphological Computation? On How the Body Contributes to Cognition and Control},
  journal      = {Artificial Life},
  year         = {2017},
  volume       = {23},
  number       = {1},
  pages        = {1--24},
  month        = {02},
  doi          = {10.1162/ARTL_a_00219}
}

@article{bartolozzi2022embodied,
  author       = {Bartolozzi, Chiara and Indiveri, Giacomo and Donati, Elisa},
  title        = {Embodied neuromorphic intelligence},
  journal      = {Nature Communications},
  year         = {2022},
  volume       = {13},
  number       = {1},
  pages        = {1024},
  doi          = {10.1038/s41467-022-28487-2}
}

@inproceedings{vanSchaik2003TauCell,
  author       = {van Schaik, A. and Jin, C.},
  title        = {The tau-cell: a new method for the implementation of arbitrary differential equations},
  booktitle    = {Proceedings of the 2003 International Symposium on Circuits and Systems, 2003. ISCAS '03.},
  year         = {2003},
  volume       = {1},
  pages        = {569--572},
  doi          = {10.1109/ISCAS.2003.1205627}
}

@inproceedings{vanSchaik2010Mihalas,
  author       = {van Schaik, André and Jin, Craig and McEwan, Alistair and Hamilton, Tara Julia and Mihalas, Stefan and Niebur, Ernst},
  title        = {A log-domain implementation of the Mihalas-Niebur neuron model},
  booktitle    = {Proceedings of 2010 IEEE International Symposium on Circuits and Systems},
  year         = {2010},
  pages        = {4249--4252},
  doi          = {10.1109/ISCAS.2010.5537563}
}

@article{bartolozzi2007synaptic,
  author       = {Bartolozzi, Chiara and Indiveri, Giacomo},
  title        = {Synaptic Dynamics in Analog VLSI},
  journal      = {Neural Computation},
  year         = {2007},
  volume       = {19},
  number       = {10},
  pages        = {2581--2603},
  doi          = {10.1162/neco.2007.19.10.2581}
}

@article{indiveri2011silicon,
  author       = {Indiveri, Giacomo and Linares-Barranco, Bernabe and Hamilton, Tara J. and van Schaik, André and Etienne-Cummings, Ralph and Delbruck, Tobi and Liu, Shih-Chii and Dudek, Piotr and Häfliger, Philipp and Renaud, Sylvie and Schemmel, Johannes and Cauwenberghs, Gert and Arthur, John and Hynna, Kai and Folowosele, Fopefolu and SAÏGHI, Sylvain and Serrano-Gotarredona, Teresa and Wijekoon, Jayawan and Wang, Yingxue and Boahen, Kwabena},
  title        = {Neuromorphic Silicon Neuron Circuits},
  journal      = {Frontiers in Neuroscience},
  year         = {2011},
  volume       = {5},
  pages        = {73},
  doi          = {10.3389/fnins.2011.00073}
}

@inproceedings{delbruck1991bump,
  author       = {Delbruck, T.},
  title        = {'Bump' circuits for computing similarity and dissimilarity of analog voltages},
  booktitle    = {IJCNN-91-Seattle International Joint Conference on Neural Networks},
  year         = {1991},
  volume       = {1},
  pages        = {475--479},
  doi          = {10.1109/IJCNN.1991.155225}
}

@inproceedings{lazzaro1989wta,
  author       = {Lazzaro, J. and Ryckebusch, S. and Mahowald, M.A. and Mead, C. A.},
  title        = {Winner-Take-All Networks of O(N) Complexity},
  booktitle    = {Advances in Neural Information Processing Systems},
  year         = {1988},
  volume       = {1},
  pages        = {703--711},
  editor       = {D. Touretzky},
  publisher    = {Morgan-Kaufmann},
  url          = {https://proceedings.neurips.cc/paper\_files/paper/1988/file/a8f15eda80c50adb0e71943adc8015cf-Paper.pdf}
}

@inproceedings{mahowald1991adaptive,
  author       = {Misha A. Mahowald},
  title        = {{Silicon retina with adaptive photoreceptors}},
  booktitle    = {Visual Information Processing: From Neurons to Chips},
  year         = {1991},
  volume       = {1473},
  pages        = {52 -- 58},
  editor       = {Bimal P. Mathur and Christof Koch},
  publisher    = {SPIE},
  organization = {International Society for Optics and Photonics},
  doi          = {10.1117/12.45540}
}

@article{lyon1988analog,
  author       = {Lyon, R.F. and Mead, C.},
  title        = {An analog electronic cochlea},
  journal      = {IEEE Transactions on Acoustics, Speech, and Signal Processing},
  year         = {1988},
  volume       = {36},
  number       = {7},
  pages        = {1119--1134},
  doi          = {10.1109/29.1639}
}

@article{xu2018carfac,
  author       = {Xu, Ying and Thakur, Chetan S. and Singh, Ram K. and Hamilton, Tara Julia and Wang, Runchun M. and van Schaik, André},
  title        = {A FPGA Implementation of the CAR-FAC Cochlear Model},
  journal      = {Frontiers in Neuroscience},
  year         = {2018},
  volume       = {12},
  pages        = {198},
  doi          = {10.3389/fnins.2018.00198}
}

@article{aono2013exploiting,
  author       = {Aono, Kenji and Shaga, Ravi K. and Chakrabartty, Shantanu},
  title        = {Exploiting Jump-Resonance Hysteresis in Silicon Auditory Front-Ends for Extracting Speaker Discriminative Formant Trajectories},
  journal      = {IEEE Transactions on Biomedical Circuits and Systems},
  year         = {2013},
  volume       = {7},
  number       = {4},
  pages        = {389--400},
  doi          = {10.1109/TBCAS.2012.2218104}
}

@article{thakur2016tab,
  author       = {Thakur, Chetan Singh and Wang, Runchun and Hamilton, Tara Julia and Tapson, Jonathan and van Schaik, André},
  title        = {A Low Power Trainable Neuromorphic Integrated Circuit That Is Tolerant to Device Mismatch},
  journal      = {IEEE Transactions on Circuits and Systems I: Regular Papers},
  year         = {2016},
  volume       = {63},
  number       = {2},
  pages        = {211--221},
  doi          = {10.1109/TCSI.2015.2512743}
}

@article{thakur2018tab,
  author       = {Thakur, Chetan Singh and Wang, Runchun and Hamilton, Tara Julia and Etienne-Cummings, Ralph and Tapson, Jonathan and van Schaik, André},
  title        = {An Analogue Neuromorphic Co-Processor That Utilizes Device Mismatch for Learning Applications},
  journal      = {IEEE Transactions on Circuits and Systems I: Regular Papers},
  year         = {2018},
  volume       = {65},
  number       = {4},
  pages        = {1174--1184},
  doi          = {10.1109/TCSI.2017.2756878}
}

@inproceedings{querlioz2013stochastic,
  author       = {Querlioz, Damien and Trauchessec, Vincent},
  title        = {Stochastic resonance in an analog current-mode neuromorphic circuit},
  booktitle    = {2013 IEEE International Symposium on Circuits and Systems (ISCAS)},
  year         = {2013},
  pages        = {1596--1599},
  doi          = {10.1109/ISCAS.2013.6572166}
}

@article{chien2018sampling,
  author       = {Chien, Chen-Han and Liu, Shih-Chii and Steimer, Andreas},
  title        = {A Neuromorphic VLSI Circuit for Spike-Based Random Sampling},
  journal      = {IEEE Transactions on Emerging Topics in Computing},
  year         = {2018},
  volume       = {6},
  number       = {1},
  pages        = {135--144},
  doi          = {10.1109/TETC.2015.2424593}
}

@article{hamilton2014stochastic,
  author       = {Hamilton, Tara Julia and Afshar, Saeed and van Schaik, André and Tapson, Jonathan},
  title        = {Stochastic Electronics: A Neuro-Inspired Design Paradigm for Integrated Circuits},
  journal      = {Proceedings of the IEEE},
  year         = {2014},
  volume       = {102},
  number       = {5},
  pages        = {843--859},
  doi          = {10.1109/JPROC.2014.2310713}
}

@article{thakur2016bayesian,
  author       = {Thakur, Chetan Singh and Afshar, Saeed and Wang, Runchun M. and Hamilton, Tara J. and Tapson, Jonathan and van Schaik, André},
  title        = {Bayesian Estimation and Inference Using Stochastic Electronics},
  journal      = {Frontiers in Neuroscience},
  year         = {2016},
  volume       = {10},
  pages        = {104},
  doi          = {10.3389/fnins.2016.00104}
}

@inproceedings{thakur2017saliency,
  author       = {Thakur, Chetan Singh and Molin, Jamal Lottier and Xiong, Tao and Zhang, Jie and Niebur, Ernst and Etienne-Cummings, Ralph},
  title        = {Neuromorphic visual saliency implementation using stochastic computation},
  booktitle    = {2017 IEEE International Symposium on Circuits and Systems (ISCAS)},
  year         = {2017},
  pages        = {1--4},
  doi          = {10.1109/ISCAS.2017.8050868}
}

@article{LyChow2010RBMFPGA,
  author  = {Ly, Daniel Le and Chow, Paul},
  title   = {High-Performance Reconfigurable Hardware Architecture for Restricted Boltzmann Machines},
  journal = {IEEE Transactions on Neural Networks},
  year    = {2010},
  volume  = {21},
  number  = {11},
  pages   = {1780--1792},
  doi     = {10.1109/TNN.2010.2073481},
  url     = {https://doi.org/10.1109/TNN.2010.2073481}
}

@article{Patel2022HardwareRBM,
  author  = {Patel, Saavan and Canoza, Philip and Salahuddin, Sayeef},
  title   = {Logically Synthesized and Hardware-Accelerated Restricted Boltzmann Machines for Combinatorial Optimization and Integer Factorization},
  journal = {Nature Electronics},
  year    = {2022},
  volume  = {5},
  number  = {2},
  pages   = {92--101},
  doi     = {10.1038/s41928-022-00714-0},
  url     = {https://doi.org/10.1038/s41928-022-00714-0}
}

@article{camsari2017pbits,
  author       = {Camsari, Kerem Yunus and Faria, Rafatul and Sutton, Brian M. and Datta, Supriyo},
  title        = {Stochastic $p$-Bits for Invertible Logic},
  journal      = {Physical Review X},
  year         = {2017},
  volume       = {7},
  number       = {3},
  pages        = {031014},
  month        = {Jul},
  doi          = {10.1103/PhysRevX.7.031014}
}

@article{cai2020intrinsic,
  author       = {Cai, Fuxi and Kumar, Suhas and Van Vaerenbergh, Thomas and Sheng, Xia and Liu, Rui and Li, Can and Liu, Zhan and Foltin, Martin and Yu, Shimeng and Xia, Qiangfei and others},
  title        = {Power-efficient combinatorial optimization using intrinsic noise in memristor Hopfield neural networks},
  journal      = {Nature Electronics},
  year         = {2020},
  volume       = {3},
  number       = {7},
  pages        = {409--418},
  doi          = {10.1038/s41928-020-0436-6}
}

@article{chen2025neurosa,
  author       = {Chen, Zihao and
 Xiao, Zhili and
 Akl, Mahmoud and
 Leugering, Johannes and
 Olajide, Omowuyi and
 Malik, Adil and
 Dennler, Nik and
 Harper, Chad and
 Bose, Subhankar and
 Gonzalez, Hector A. and
 Samaali, Mohamed and
 Liu, Gengting and
 Eshraghian, Jason and
 Pignari, Riccardo and
 Urgese, Gianvito and
 Andreou, Andreas G. and
 Shankar, Sadasivan and
 Mayr, Christian and
 Cauwenberghs, Gert and
 Chakrabartty, Shantanu},
  title        = {ON-OFF neuromorphic ISING machines using Fowler-Nordheim annealers},
  journal      = {Nature Communications},
  year         = {2025},
  volume       = {16},
  number       = {1},
  pages        = {3086},
  doi          = {10.1038/s41467-025-58231-5}
}

@article{ahsan2026higherorder,
  author       = {Ahsan, Faiek and
 Maiti, Saptarshi and
 Chen, Zihao and
 Kaiser, Jakob and
 Nandi, Ankita and
 Srivatsav, Madhuvanthi and
 Schemmel, Johannes and
 Andreou, Andreas G. and
 Eshraghian, Jason and
 Thakur, Chetan Singh and
 Chakrabartty, Shantanu},
  title        = {Higher-order neuromorphic Ising machines---autoencoders and Fowler-Nordheim annealers are all you need for scalability},
  journal      = {Nature Communications},
  year         = {2026},
  volume       = {17},
  number       = {1},
  pages        = {5293},
  doi          = {10.1038/s41467-026-71937-4}
}

@article{qiao2015rolls,
  author       = {Qiao, Ning and Mostafa, Hesham and Corradi, Federico and Osswald, Marc and Stefanini, Fabio and Sumislawska, Dora and Indiveri, Giacomo},
  title        = {A reconfigurable on-line learning spiking neuromorphic processor comprising 256 neurons and 128K synapses},
  journal      = {Frontiers in Neuroscience},
  year         = {2015},
  volume       = {9},
  pages        = {141},
  doi          = {10.3389/fnins.2015.00141}
}

@article{Moradi2018DYNAPs,
  author       = {Moradi, Saber and Qiao, Ning and Stefanini, Fabio and Indiveri, Giacomo},
  title        = {A Scalable Multicore Architecture With Heterogeneous Memory Structures for Dynamic Neuromorphic Asynchronous Processors (DYNAPs)},
  journal      = {IEEE Transactions on Biomedical Circuits and Systems},
  year         = {2018},
  volume       = {12},
  number       = {1},
  pages        = {106--122},
  doi          = {10.1109/TBCAS.2017.2759700}
}

@article{wang2018fpga,
  author       = {Wang, Runchun M. and Thakur, Chetan S. and van Schaik, André},
  title        = {An FPGA-Based Massively Parallel Neuromorphic Cortex Simulator},
  journal      = {Frontiers in Neuroscience},
  year         = {2018},
  volume       = {12},
  pages        = {213},
  doi          = {10.3389/fnins.2018.00213}
}

@article{boahen2000aer,
  author       = {Boahen, K.A.},
  title        = {Point-to-point connectivity between neuromorphic chips using address events},
  journal      = {IEEE Transactions on Circuits and Systems II: Analog and Digital Signal Processing},
  year         = {2000},
  volume       = {47},
  number       = {5},
  pages        = {416--434},
  doi          = {10.1109/82.842110}
}

@inproceedings{imam2012qdi,
  author       = {Imam, Nabil and Akopyan, Filipp and Arthur, John and Merolla, Paul and Manohar, Rajit and Modha, Dharmendra S.},
  title        = {A Digital Neurosynaptic Core Using Event-Driven QDI Circuits},
  booktitle    = {2012 IEEE 18th International Symposium on Asynchronous Circuits and Systems},
  year         = {2012},
  pages        = {25--32},
  doi          = {10.1109/ASYNC.2012.12}
}

@article{li2025neuroscale,
  author       = {Li, Congyang and Imam, Nabil and Manohar, Rajit},
  title        = {A deterministic neuromorphic architecture with scalable time synchronization},
  journal      = {Nature Communications},
  year         = {2025},
  volume       = {16},
  number       = {1},
  pages        = {10329},
  doi          = {10.1038/s41467-025-65268-z}
}

@article{merolla2014truenorth,
  author       = {Paul A. Merolla and John V. Arthur and Rodrigo Alvarez-Icaza and Andrew S. Cassidy and Jun Sawada and Filipp Akopyan and Bryan L. Jackson and Nabil Imam and Chen Guo and Yutaka Nakamura and Bernard Brezzo and Ivan Vo and Steven K. Esser and Rathinakumar Appuswamy and Brian Taba and Arnon Amir and Myron D. Flickner and William P. Risk and Rajit Manohar and Dharmendra S. Modha},
  title        = {A million spiking-neuron integrated circuit with a scalable communication network and interface},
  journal      = {Science},
  year         = {2014},
  volume       = {345},
  number       = {6197},
  pages        = {668--673},
  doi          = {10.1126/science.1254642}
}

@article{diorio1996synapse,
  author       = {Diorio, C. and Hasler, P. and Minch, A. and Mead, C.A.},
  title        = {A single-transistor silicon synapse},
  journal      = {IEEE Transactions on Electron Devices},
  year         = {1996},
  volume       = {43},
  number       = {11},
  pages        = {1972--1980},
  doi          = {10.1109/16.543035}
}

@article{indiveri2006vlsi,
  author       = {Indiveri, G. and Chicca, E. and Douglas, R.},
  title        = {A VLSI array of low-power spiking neurons and bistable synapses with spike-timing dependent plasticity},
  journal      = {IEEE Transactions on Neural Networks},
  year         = {2006},
  volume       = {17},
  number       = {1},
  pages        = {211--221},
  doi          = {10.1109/TNN.2005.860850}
}

@article{friedmann2017hybrid,
  author       = {Friedmann, Simon and Schemmel, Johannes and Grübl, Andreas and Hartel, Andreas and Hock, Matthias and Meier, Karlheinz},
  title        = {Demonstrating Hybrid Learning in a Flexible Neuromorphic Hardware System},
  journal      = {IEEE Transactions on Biomedical Circuits and Systems},
  year         = {2017},
  volume       = {11},
  number       = {1},
  pages        = {128--142},
  doi          = {10.1109/TBCAS.2016.2579164}
}

@inproceedings{wang2016stochasticSTDP,
  author       = {Wang, Runchun and Thakur, Chetan Singh and Hamilton, Tara Julia and Tapson, Jonathan and van Schaik, André},
  title        = {A stochastic approach to STDP},
  booktitle    = {2016 IEEE International Symposium on Circuits and Systems (ISCAS)},
  year         = {2016},
  pages        = {2082--2085},
  doi          = {10.1109/ISCAS.2016.7538989}
}

@article{niazi2024boltzmann,
  author       = {Niazi, Shaila and Chowdhury, Shuvro and Aadit, Navid Anjum and Mohseni, Masoud and Qin, Yao and Camsari, Kerem Y},
  title        = {Training deep Boltzmann networks with sparse Ising machines},
  journal      = {Nature Electronics},
  year         = {2024},
  volume       = {7},
  number       = {7},
  pages        = {610--619},
  doi          = {10.1038/s41928-024-01182-4}
}

@article{milde2017obstacle,
  author       = {Milde, Moritz B. and Blum, Hermann and Dietmüller, Alexander and Sumislawska, Dora and Conradt, Jörg and Indiveri, Giacomo and Sandamirskaya, Yulia},
  title        = {Obstacle Avoidance and Target Acquisition for Robot Navigation Using a Mixed Signal Analog/Digital Neuromorphic Processing System},
  journal      = {Frontiers in Neurorobotics},
  year         = {2017},
  volume       = {11},
  pages        = {28},
  doi          = {10.3389/fnbot.2017.00028}
}

@inproceedings{glatz2019adaptive,
  author       = {Glatz, Sebastian and Martel, Julien and Kreiser, Raphaela and Qiao, Ning and Sandamirskaya, Yulia},
  title        = {Adaptive motor control and learning in a spiking neural network realised on a mixed-signal neuromorphic processor},
  booktitle    = {2019 International Conference on Robotics and Automation (ICRA)},
  year         = {2019},
  pages        = {9631--9637},
  doi          = {10.1109/ICRA.2019.8794145}
}

@article{yik2025neurobench,
  author       = {Yik, Jason and Van den Berghe, Korneel and Den Blanken, Douwe and Bouhadjar, Younes and Fabre, Maxime and Hueber, Paul and Ke, Weijie and Khoei, Mina A and Kleyko, Denis and Pacik-Nelson, Noah and others},
  title        = {The neurobench framework for benchmarking neuromorphic computing algorithms and systems},
  journal      = {Nature Communications},
  year         = {2025},
  volume       = {16},
  number       = {1},
  pages        = {1545},
  doi          = {10.1038/s41467-025-56739-4}
}

\end{document}